\documentclass[letterpaper, 10 pt, conference]{ieeeconf}  %

\IEEEoverridecommandlockouts                              %

\usepackage{graphics} %
\usepackage{epsfig} %
\usepackage{mathptmx} %
\usepackage{times} %
\usepackage{amsmath} %
\usepackage{amssymb}  %
\usepackage{tabularx}
\usepackage{makecell}
\usepackage{caption}
\usepackage{multirow}
\usepackage{ulem}
\usepackage[linkcolor=black,citecolor=black,urlcolor=black,colorlinks=true]{hyperref}

\title{\LARGE \bf
EasyHeC++: Fully Automatic Hand-Eye Calibration \\ with Pretrained Image Models
}

\author{
    \quad Zhengdong Hong$^{1*}$
    \quad Kangfu Zheng$^{2*}$
    \quad Linghao Chen$^{1}$
    \\
    $^1$ Zhejiang University \quad
    $^2$ Tsinghua University \quad
    \thanks{
        $^*$Equal contribution.
        Corresponding author: Linghao Chen.}%
}

\begin{document}

    \maketitle
    \thispagestyle{empty}
    \pagestyle{empty}

    \begin{abstract}
    Hand-eye calibration plays a fundamental role in robotics by directly influencing the efficiency of critical operations such as manipulation and grasping.
    In this work, we present a novel framework, EasyHeC++, designed for fully automatic hand-eye calibration.
    In contrast to previous methods that necessitate manual calibration, specialized markers, or the training of arm-specific neural networks, our approach is the first system that enables accurate calibration of any robot arm in a marker-free, training-free, and fully automatic manner.
    Our approach employs a two-step process.
    First, we initialize the camera pose using a sampling or feature-matching-based method with the aid of pretrained image models.
    Subsequently, we perform pose optimization through differentiable rendering.
    Extensive experiments demonstrate the system's superior accuracy in both synthetic and real-world datasets across various robot arms and camera settings.
    Project page: \url{https://ootts.github.io/easyhec_plus/}.
\end{abstract}

    \section{Introduction}
\label{sec:intro}

Hand-eye calibration is a fundamental problem in robotics.
It connects the vision system and the robot arm system by transforming the perception of the camera into the robot's coordinate system.
This is crucial for many robotic applications, such as robotic grasping~\cite{miller2004graspit,chen2023perceiving}, robotic manipulation~\cite{jia2023chain, an2023rgbmanip}, and robotic assembly~\cite{zobov2023auto}.

Traditionally, the hand-eye calibration problem is addressed by using a marker~\cite{garrido2014automatic,ilonen2011robust,tsai1989new} to assist computing the camera pose by solving a $AX=XB$ or $AX=YB$ equation~\cite{ilonen2011robust,tsai1989new,park1994robot,daniilidis1999hand,Horaud1995Hec}.
These methods not only necessitate the placement of a high-quality marker in the scene but also require the manual selection of a series of joint poses.
This manual process is time-consuming, and not user-friendly,  thereby restricting their applicability in real-world lab and household scenarios.

\begin{figure}
    \centering
    \resizebox{\linewidth}{!}{
        \includegraphics[width=0.5\linewidth,clip]{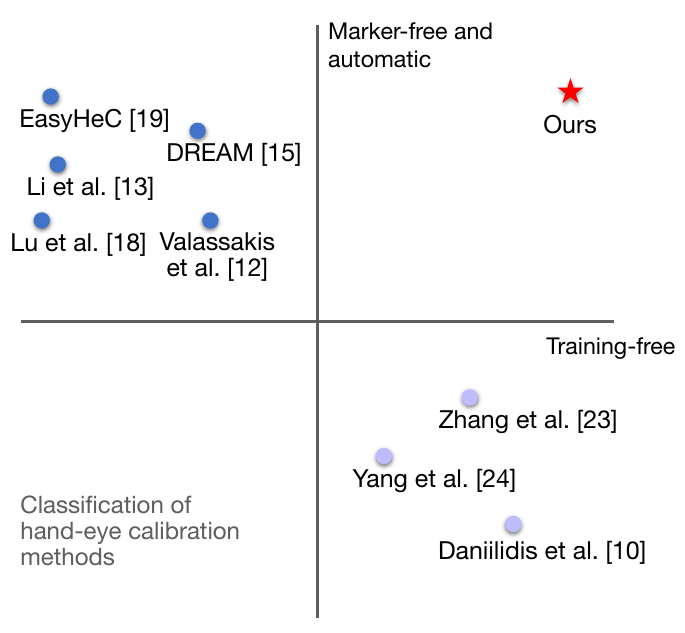}
    }
    \caption{\textbf{Comparison between our method and previous methods. }
    Our method not only delivers high accuracy but also is fully automatic, marker-free, and training-free.
    }
    \vspace{-1em}
    \label{fig:teaser}
\end{figure}

Recently, learning-based methods have been proposed to address the hand-eye calibration problem.
These methods typically involve employing a neural network to either directly regress the camera pose~\cite{valassakis2022learning,li2023look} or detect keypoints of the robot arm~\cite{bahadir2022deep,lee2020dream,lu2022pose}, followed by solving the camera pose using the Perspective-n-Point (PnP) algorithm~\cite{lepetit2009epnp}.
The performance of these methods is limited by the quality and quantity of the training data.
Moreover, one trained model can only be applied to a single type of robot arm.
If the need arises to calibrate a different type of robot arm, it requires starting the process again, which involves collecting new data and training a new model.

Recent works~\cite{lu2023image,easyhec} propose to use differentiable rendering to optimize the camera pose with a pixel-wise mask loss and deliver superior accuracy.
These methods get rid of accurate marker placement and intricately selected joint poses, thus highly suitable in household scenarios, especially when re-calibration is needed at deployment time. However, they still require a rough pose as initialization and a mask as supervision.
The mask segmentation network requires training for each type of robot arm.
In addition, the pose initialization is either manually set or obtained by another neural network. Both factors have increased human efforts and training costs, prohibiting these methods from being fully automatic and widely adopted.

In this paper, we introduce EasyHeC++ offering the following features:
\begin{itemize}
    \item Accurate (3mm error) and fully automatic hand-eye calibration system.
    \item The first method without any training cost or manual annotation for any type of robot arm.
    \item Fast setup, 15 minutes for a new robot arm, and 5 minutes for re-calibration.
    \item Both eye-to-hand and eye-in-hand configurations are supported.
\end{itemize}
Our core idea is to integrate the outstanding generalization abilities of pre-trained image models with the precision derived from the differentiable-rendering-based pose optimization.
This innovative design simplifies the prerequisites by requiring only a real robot arm, a camera, and the robot arm model.
Consequently, this approach significantly diminishes human intervention and training costs, rendering our method more practical for real-world applications.

Our framework consists of two main components: 1) pose initialization and 2) pose optimization.
In the pose initialization phase, we utilize either a sampling-based method or a feature-matching-based method to initialize the camera pose for the initial calibration and subsequent re-calibrations, respectively.
The sampling-based initialization selects a pose from a set of densely sampled camera poses on a hemisphere.
We then refine this pose using differentiable rendering under the supervision of the masks generated by a text-prompted segmentation model~\cite{liu2023grounding}.
On the other hand, the feature-matching-based initialization leverages the historical calibrated image-pose pairs to infer an initial camera pose.
These methods, in contrast to training a neural network or manually setting the camera pose, provide a more automatic approach and are not constrained by the type of robot arm.
Next, in the pose optimization phase, we follow the methods outlined in ~\cite{easyhec}, which optimizes the camera pose with a pixel-wise mask loss and uses a space exploration module to search for the most informative joint pose, yielding more precise calibration results.
In this phase, the mask supervision is generated by a pre-trained open-world segmentation model~\cite{kirillov2023segany}, using the projected robot arm model as prompts.
Leveraging the kinematics model of the robot arm to guide the pretrained image models fully capitalizes on their generalization ability and allows us to generate an accurate mask without the need for any manual prompt annotation or training tailored to a specific type of robot arm.
After the camera pose is solved, we incorporate the calibrated image-pose pair into the database.
This information will serve as the reference for pose initialization in subsequent re-calibrations using the feature-matching-based method.

We evaluate the proposed method across various types of robot arms in synthetic and real-world scenes.
The results demonstrate that our method outperforms all the previous methods in terms of accuracy and automation under both eye-to-hand and eye-in-hand settings.
To the best of our knowledge, this is the first work that achieves fully automatic hand-eye calibration, eliminating the necessity for training, manual annotation, or markers.
In a commitment to contribute to the robotics community, we are dedicated to open-sourcing our system for wider accessibility and benefit.

    \section{Related Work}

\subsection{Hand-Eye Calibration}

Hand-eye calibration is the problem of obtaining the transformation matrix between the camera and the robot reference frames, which involves two camera settings: eye-to-hand calibration and eye-in-hand calibration.
In eye-to-hand calibration, the camera is stationary relative to the robot base.
Conversely, for eye-in-hand calibration, the camera is stationary relative to the robot's end-effector.

\medskip\noindent\textbf{Traditional methods.}
Traditional methods employ a marker to solve AX=XB hand-eye formulation for eye-to-hand calibration~\cite{ilonen2011robust,tsai1989new,park1994robot,daniilidis1999hand} or AX=YB robot-world-hand-eye formulation for eye-in-hand calibration~\cite{zhuang1994simultaneous}.
Several recent approaches have integrated these formulations with active next-best-view selection to make the calibration process more automated and precise.
Zhang et al.~\cite{zhang2023active} proposed an online estimated discrete viewing quality field to represent the calibration quality of selected camera views.
Yang et al.~\cite{yang2023next} introduced uncertainty reduction to guide the robot pose selection.
However, these approaches are highly affected by the visibility and quality of the markers, which limits their applicability and increases the difficulty for users to utilize them.

\medskip\noindent\textbf{Learning-based methods.}
Recent works proposed learning-based or marker-free methods to solve hand-eye calibration.
For eye-to-hand setting, DREAM~\cite{lee2020dream} introduced a two-step framework, which first utilizes a deep neural network to detect 2D projections of keypoints from the RGB image of the robot, and then recovers the camera poses using Perspective-n-Point (PnP) algorithm~\cite{lepetit2009epnp}.
Similarly, Lu et al.~\cite{lu2022pose} proposed to first find the optimal set of keypoints on the robot through an iterative approach using DNNs, and subsequently to estimate pose utilizing the PnP algorithm.

\medskip\noindent\textbf{Optimization-based methods.}
Recently, Lu et al.~\cite{lu2023image} proposed to use differentiable rendering to optimize the camera pose with a pixel-wise mask loss as the objective function.
EasyHeC~\cite{easyhec} further designed an uncertainty-based space exploration module to search for the next best informative joint pose for more accurate calibration results for the eye-to-hand setting.
For the eye-in-hand setting, Valassakis et al.~\cite{valassakis2022learning} trained a simple neural network and directly regressed the camera pose from images captured by a mounted camera.
For both eye-to-hand and eye-in-hand settings, Li et al.~\cite{li2023look} proposed to detect the robot base and align it with a point cloud to compute the camera pose.
All the existing methods either necessitate manual calibration with specialized markers or require the training of arm-specific neural networks.
As a comparison, we are the first hand-eye-calibration system in a marker-free, training-free, and fully automatic manner.

\medskip\noindent\textbf{Difference from EasyHeC}
Our method is built upon EasyHeC, however, there are several key differences.
\textbf{(1)} EasyHeC requires mask and pose networks re-training for each new type of robot arm, while our method is training-free and can be applied to any robot arm.
\textbf{(2)} EasyHeC is designed only for one-time calibration, while we support fast recalibration after a camera repositioning.
\textbf{(3)} EasyHeC only supports the eye-to-hand setting, while we also support the eye-in-hand setting.

\subsection{Visual Localization}
Visual localization is an important computer vision task that aims to estimate the camera pose of a new image given a known scene representation.
The scene is usually represented as the reconstruction results of Structure-from-Motion~\cite{schonberger2016structure} using feature-matching-based methods~\cite{detone2018superpoint,sarlin2020superglue}.
Then the core problem of visual localization becomes finding the correspondences between the pixels in the 2D image and the 3D points in the scene.
HLoc~\cite{sarlin2019coarse} proposed to address the visual localization problem by image retrieval and lifting 2D-2D matchings to 2D-3D matchings in a coarse-to-fine manner.
OnePose~\cite{sun2022onepose} proposed to first reconstruct a semi-dense point cloud representation for the 3D object and then train a neural network to directly match the 2D pixel to the 3D point cloud.
OnePose++~\cite{he2022onepose++} proposed to substitute the COLMAP~\cite{schonberger2016structure} with a learning-based feature matching approach~\cite{sun2021loftr} to improve the performance on textureless objects.
However, these methods are not suitable for our scenarios.
The reconstruction in their methods has no canonical space, whereas the robot arm has a pre-defined canonical space.
Moreover, the robot arm has a known object shape, making aligning them highly reliant on the reconstructed shape quality and camera pose accuracy.
The most relevant work in this field is MeshLoc~\cite{panek2022meshloc}, which proposed to use a mesh-based representation to avoid feature matching between database images.
In terms of hand-eye calibration, the robot mesh is usually off-the-shelf and can be easily obtained.

\subsection{Image Models for Segmentation}

Traditional learning-based methods using networks like~\cite{he2017mask,kirillov2020pointrend} are largely confined by their generalizability in tasks and data distributions beyond those seen during training.
In our task, as we aim to segment the highly-articulated robotic arm, those models require us to collect extra robot arm images for training, which is laborious.
The current trend in large vision language models like Segment Anything (SAM)~\cite{kirillov2023segany} has enabled precise zero-shot image segmentation.
Boosted by its scaled model size and abundant text corpora from the web, SAM~\cite{kirillov2023segany} and its follow-up works~\cite{kirillov2023segany, mobile_sam, sam_hq, li2023semantic} dominate image segmentation by its superiority in quality, speed, and generalizability.
Another virtue of large vision language models is they are designed and trained to be promotable. Grounded-Segment-Anything~\cite{liu2023grounding} further broadens the scope of the application by supporting image, text, and speech inputs.
In our pipeline, we use the kinematics model of the robotic arm to guide the SAM model~\cite{liu2023grounding} to generate masks without the need for manually labeled prompts.

    \section{Methods}

\subsection{Background}
In this work, we aim to address the hand-eye calibration problem under two settings: eye-to-hand calibration and eye-in-hand calibration. We represent the relative pose between the camera and the robot base as $T_{cb}$ and the relative pose between the camera and the end-effector as $T_{ce}$.

Our method is built on EasyHeC, which addresses the eye-to-hand calibration problem iteratively.
Each iteration includes two main components: differentiable-rendering-based camera pose optimization and consistency-based joint space exploration.
The differentiable-rendering-based optimization uses a pixel-wise rendering mask loss to optimize the camera pose $T_{cb}$ as follows:
\begin{equation}
    L\left(\xi_{c b}\right)=\left(\min \left(1, \sum_l \pi\left(\exp \left(\xi_{c b}\right) T_{b l} l\right)\right)-M\right)^2,\label{eq:equation_dr}
\end{equation}
where $\xi_{cb} \in \mathfrak{se}(3)$ is the exponential coordinate of the relative pose between the camera and the robot base, $\pi$ is a differentiable mask renderer, $T_{bl}$ is the relative pose between the base link and the link $l$, computed from forward kinematics, and $M$ is the observed mask, inferred from the RGB image captured by the camera $c$.

After the optimization of each iteration, the consistency-based joint space exploration is performed.
This process samples a bunch of joint poses in the simulator and identifies the most informative one to improve the accuracy of the calibration results.
Then, the robot arm moves to the next joint pose and the optimization process is performed again on all the collected images.
This process is repeated until the number of iterations reaches a pre-defined maximum number.
We call this number of iterations as the space exploration iterations in the following sections. To learn more details, please refer to~\cite{easyhec}.

\subsection{Overview}
\begin{figure*}[ht]
    \centering
    \includegraphics[width=\linewidth]{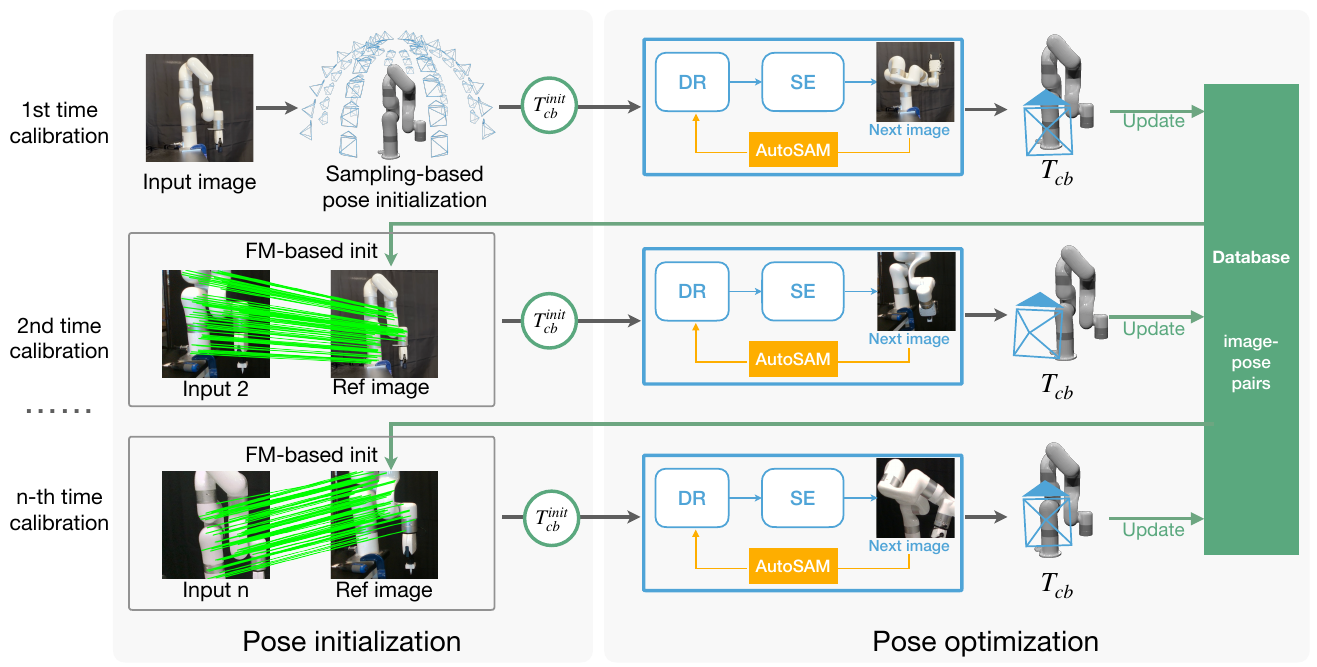}
    \caption{
        \textbf{EasyHeC++ architecture.}
        We consider not only single-instance calibration but also recalibration after camera movement.
        At each time of calibration, EasyHeC++ consists of two main components: pose initialization and pose optimization.
        At the first time of calibration, we use a sampling-based pose initialization module to initialize a rough camera pose ${T_{cb}^{init}}$,
        while in subsequent re-calibrations, we use a feature matching (FM)-based module to initialize the camera pose, using the historical image-pose pairs in the database.
        Then we run pose optimization by first using a differentiable renderer (DR) to optimize the camera pose and then running a space exploration (SE) to obtain the next joint pose to increase the accuracy.
        In this process, AutoSAM is proposed to predict mask as the supervision to the DR process.
        After solving the camera pose $T_{cb}$, we add the image-pose pair to the database.
    }
     \vspace{-1.5em}
    \label{fig:pipeline}
\end{figure*}

As shown in Fig.~\ref{fig:pipeline}, EasyHeC++ aims to solve the hand-eye calibration problem in a fully automatic manner.
Notably, we address not only the first-time calibration but also the subsequent re-calibration after the camera movement.
At each time of calibration, we first initialize the camera pose either via a sampling-based method (Sec.~\ref{sampling-based-pose-init}) or a feature-matching-based method using the existing image-pose pairs in the database (Sec.~\ref{fm-based-pose-init}).
Next, we perform the pose optimization in the same way as EasyHeC, using differentiable-rendering-based optimization and consistency-based joint space exploration.
The main difference is that we design an AutoSAM module to automatically predict segmentation masks (Sec.~\ref{auto-sam}) as the supervision.
After each time of calibration, the image-pose pair is added to the database as the reference images for the initialization in the subsequent re-calibrations.
We use eye-to-hand calibration as the default setting in Sec.~\ref{auto-pose-init} and Sec.~\ref{auto-sam} and finally discuss the eye-in-hand calibration in Sec.~\ref{in-hand-optimization}.

\subsection{Automatic Pose Initialization}\label{auto-pose-init}
The hand-eye-calibration accuracy is highly dependent on the quality of the pose initialization.
In the previous work~\cite{easyhec}, the pose initialization is obtained by a neural network.
The network has to be trained on synthetic data and fine-tuned on real data for each type of robot arm, which may suffer from the sim-to-real domain gap and not be user-friendly due to the additional effort of per-arm data annotation and training.
In this work, we aim to initialize the pose in an automatic and training-free manner.

\subsubsection{Sampling-based Pose Initialization}\label{sampling-based-pose-init}

Given a robot arm that has never been calibrated before, we aim to estimate an initial camera pose $T_{cb}^{init}$ for the following differentiable-rendering-based pose optimization process.

The basic idea is to generate segmentation masks using the large segmentation model, GroundedSAM~\cite{liu2023grounding}, with ``robot arm'' as the text prompt and enumerate the camera poses to find the one with the highest similarity to the observed mask.
Concretely, we sample a bunch of camera poses on a hemisphere as in Fig.~\ref{fig:pipeline}.
Then we render the robot arm model to obtain the rendered mask and compute the mask IoU between the rendered and observed masks for each camera pose.
Then we select the camera pose with the highest mask IoU and further refine it using differentiable rendering in Eq.~\ref{eq:equation_dr} to obtain the final initial pose $T_{cb}^{init}$, where the mask predicted by GroundedSAM is used to compute the mask loss.

In practice, the process of sampling different joint poses is only required for the first time.
In the subsequent re-calibrations, we adopt the feature-matching-based initialization introduced in the next subsection, which only requires a single joint pose as input.

\subsubsection{Feature-Matching-based Pose Initialization}\label{fm-based-pose-init}
As shown in Fig.~\ref{fig:pipeline}, after each time of calibration, we update the database with the image at the initial joint pose and the solved camera pose.
When re-calibration is needed due to the camera movement, we utilize the existing image-pose pairs to initialize the camera pose via a feature-matching technique as in~\cite{panek2022meshloc}.

Specifically, for an image-pose pair $\{I^r, T_{cb}^r\}$ in the database, we can obtain $\{p^r,P^c\}$ as the correspondences between the 2D pixels and the 3D point cloud in the robot arm canonical space.
Then, given the target image $I^t$, we can adopt a pre-trained dense feature-matching network to predict the pixel correspondences $\{p^r, p^t\}$ between the reference image $I^r$ and the target image $I^t$.
Then the 2D-2D correspondences are lifted to 2D-3D correspondences $\{p^t,P^c\}$ to solve the camera pose $T_{cb}^t$.

Benefiting from the nature of the feature-matching network which operates on local patches, this method is not restricted by the type of robot arm and is thus fully automatic.

\subsection{Automatic Segmentation Prediction}\label{auto-sam}

Predicting an accurate segmentation mask of the robot arm plays a crucial role in the differentiable-rendering-based optimization process since its quality directly impacts the resulting calibration accuracy.
Previous methods~\cite{lu2023image,easyhec} train neural networks~\cite{chen2017rethinking,kirillov2020pointrend} to predict this segmentation mask, which is not only time-consuming but also restricted to a specific type of robot arm.
The recent large vision language models~\cite{sam_hq,li2023semantic,kirillov2023segany,liu2023grounding} have shown impressive performance in the open-world image segmentation task, but they either require a manual prompt as input~\cite{sam_hq,kirillov2023segany} or deliver inferior mask predictions~\cite{liu2023grounding}.

In this work, we design a module called AutoSAM, which uses the kinematics model of the robot arm as a guide to SAM~\cite{kirillov2023segany} to generate the masks in a fully automatic and training-free manner.
The basic idea is to use the initialized camera pose or the optimized camera pose in the last space exploration iteration to project the robot arm model to the current frame.
Then the projection is used as a prompt to generate the segmentation mask as $M_l = \text{SAM} (I_t, \Pi (q_t,l; T_{cb}^{t-1})),$
where $M_l$ is the predicted mask of link $l$, $I_t$ is the RGB image at the current joint pose $q_t$, $\Pi$ is a projection operation producing a 2D bounding box for link $l$, and $T_{cb}^{t-1}$ is the camera pose solved in the last space exploration iteration or the initialized camera pose in the first iteration.
Then, we combine all the masks of the links to obtain the final mask $M$.
Practically, in addition to link-wise masks, we also use the 2D bounding box of every 2 adjacent links as the prompt to further improve the mask quality.
As shown in Sec.~\ref{sec:ablation_study_sam}, this link-wise prompted SAM can generate superior mask quality than the previous methods while requiring zero training cost.

\subsection{Eye-In-Hand Camera Pose Optimization}\label{in-hand-optimization}

Eye-in-hand calibration is another setting under the hand-eye calibration problem, in addition to the to-hand setting.
Different from previous eye-in-hand works that rely on a marker~\cite{zhang2023active,yang2023next} or use a neural network to predict the camera pose~\cite{valassakis2022learning}, we follow the previous work, EasyHeC~\cite{easyhec}, which uses a pixel-wise mask loss to optimize the camera pose.
Specifically, the loss function is defined as follows:
\begin{equation}
    L(T_{ce})=\left(\min \left(1, \sum_l \pi\left(\exp \left(\xi_{c e}\right) T_{eb} T_{bl} l\right)\right)-M\right)^2,
\end{equation}
where $ T_{eb}$ and $T_{bl}$ are the relative pose between the end-effector and the robot base and the relative pose between the robot base and the link $l$, respectively, computed from forward kinematics, $\xi_{ce}$ is exponential coordinate of the relative pose between the camera and the end-effector,
and $M$ is the observed mask generated by AutoSAM.

\begin{figure}[ht]
    \centering
    \resizebox{1.0 \columnwidth}{!}{
        \begin{tabular}{cc}
            \includegraphics[width=0.5\linewidth,trim={0cm 0cm 0cm 0cm},clip]{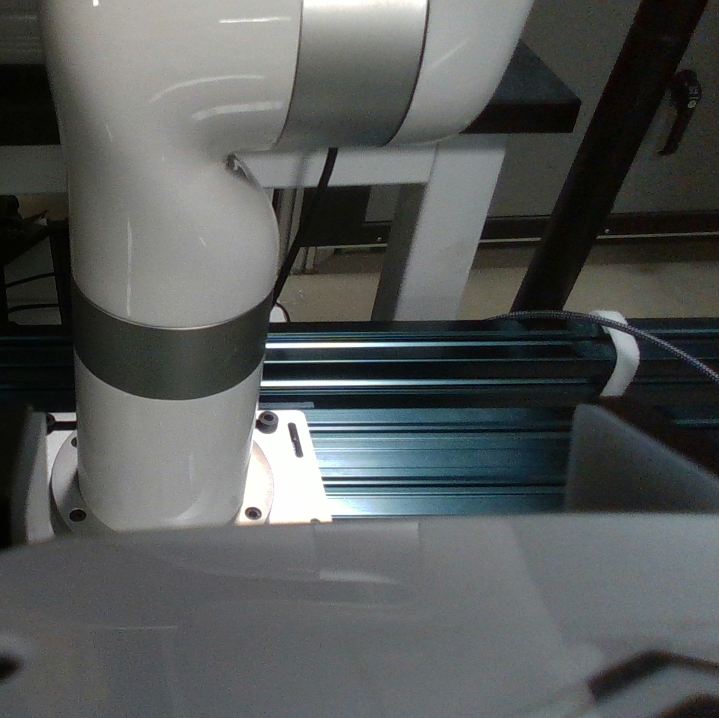} &
            \includegraphics[width=0.5\linewidth,trim={0cm 0cm 0cm 0cm},clip]{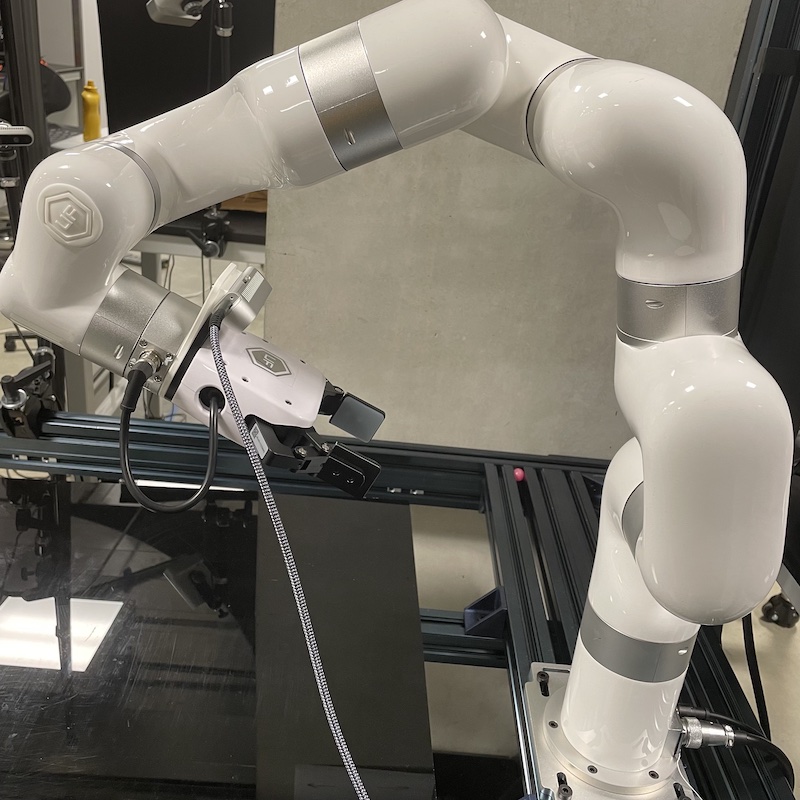}\\
            (a) & (b) \\
        \end{tabular}
    }
    \caption{\textbf{Example images for our method under the eye-in-hand setting.}
    (a) is the image captured by the in-hand camera and (b) is the image captured from the spectator's view for illustration.
    }
    \vspace{-1em}
    \label{fig:in_hand_example}
\end{figure}

Because of the nature of the eye-in-hand setting, it is not guaranteed that the robot arm is visible in the image at a random joint pose.
Thus we propose to use a different joint pose sampling strategy from the eye-to-hand setting to ensure the robot arm is visible in the image.
Specifically, instead of sampling the angle of each joint independently as in~\cite{easyhec}, we first sample the end-effector pose to ensure the end-effector is oriented towards the robot base link properly, and then we compute the joint poses using inverse kinematics.
An example image is shown in Fig.~\ref{fig:in_hand_example}.

Moreover, we observe that the gripper often occupies a large region in the image.
This region is usually not informative for camera pose optimization.
Besides, this part is an edge area with low undistortion quality in the image, so including this part in optimization could lead to a significant accuracy drop.
To address this problem, we propose to ignore this region in the optimization process with the following mask loss:
\begin{equation}
    L(T_{ce})=\sum_{p \notin O}\left(\min \left(1, \sum_l \pi\left(\exp \left(\xi_{c e}\right) T_{eb} T_{bl} l\right)\right)-M\right)_p^2,
\end{equation}

where $p$ and $O$ are a pixel and the region of the gripper in the image, respectively.

\subsection{Implementation Details}
In the sampling-based pose initialization module, the distance between the camera and the robot arm is fixed to 1 meter, and the elevation angle is sampled from 0 degrees to 70 degrees with an interval of 10 degrees.
The azimuth angle is sampled from 0 degrees to 360 degrees with an interval of 30 degrees.
In the feature-matching-based pose initialization module, we first use image retrieval~\cite{arandjelovic2016netvlad} to find the image that is the most similar to the target image and use the feature-matching-based method to initialize the pose.
we use DKM~\cite{edstedt2023dkm} as our feature-matching network since it handles texture-less and specular robot arm surfaces quite well.
Then we use PnP with RANSAC to solve the pose of the target image based on the correspondences.
Other hyperparameters in the differentiable-rendering-based pose optimization and consistency-based joint space exploration are the same as EasyHeC~\cite{easyhec}.
For the initial calibration in the eye-in-hand setting, instead of using the sampling-based pose initialization module, we manually initialize the camera pose since this pose usually demonstrates much smaller variation compared to the eye-to-hand setting.
Other operations remain the same as in the eye-to-hand setting.

    \section{Experiments}

\subsection{Evaluation on Synthetic Datasets}

\subsubsection{Eye-to-Hand Setting}

In this section, we evaluate our method using a synthetic dataset proposed in~\cite{easyhec}.
The dataset consists of 100 scenes captured under different camera poses under the eye-to-hand setting.
We conduct a comparative analysis, evaluating the performance of our method in comparison to previous approaches.
Following the evaluation protocols in~\cite{easyhec}, we only evaluate the scenes where~\cite{tsai1989new} and~\cite{lee2020dream} successfully solve the camera pose.
\begin{table}[ht]
    \centering
    \renewcommand{\arraystretch}{1.2}
    \resizebox{\linewidth}{!}{
        \begin{tabular}{c|c|ccccc}
            \hline
            Method & \#views & \multicolumn{5}{c}{Rotation error ($^\circ$)} \\
            \Xhline{3\arrayrulewidth}
            Marker-based~\cite{tsai1989new} & 20 &  \multicolumn{5}{c}{ 0.870} \\
            \hline
            DREAM~\cite{lee2020dream}  & 1 $\sim$ 5     & 1.924          & 1.240          & 0.981          & 0.764          & 0.704          \\
            \hline
            EasyHeC~\cite{easyhec}  & 1 $\sim$ 5           & 0.322          & 0.128          & 0.109          & 0.097          & 0.081          \\
            \hline
            Ours                 & 1 $\sim$ 5              & \textbf{0.246} & \textbf{0.076} & \textbf{0.064} & \textbf{0.058} & \textbf{0.045} \\
            \hline
            \Xhline{3\arrayrulewidth}
        \end{tabular}
    }
    \caption{
        \textbf{Rotation error evaluation results on the xArm synthetic dataset.}
    }
    \vspace{-1em}
    \label{tab:xarm_synthetic_rotation}
\end{table}

\begin{table}[ht]
    \centering
    \renewcommand{\arraystretch}{1.2}
    \resizebox{\linewidth}{!}{

        \begin{tabular}{c|c|ccccc}
            \hline
            Method & \#views & \multicolumn{5}{c}{Translation error (cm)} \\
            \Xhline{3\arrayrulewidth}
            Marker-based & 20         &   \multicolumn{5}{c}{2.000 } \\
            \hline
            DREAM        & 1 $\sim$ 5 & 0.529          & 0.473          & 0.374          & 0.347          & 0.303          \\
            \hline
            EasyHeC      & 1 $\sim$ 5 & 0.488          & 0.298          & 0.252          & 0.206          & 0.206          \\
            \hline
            Ours         & 1 $\sim$ 5 & \textbf{0.318} & \textbf{0.176} & \textbf{0.159} & \textbf{0.137} & \textbf{0.135} \\
            \Xhline{3\arrayrulewidth}
        \end{tabular}
    }
    \caption{
        \textbf{Translation error evaluation on the xArm synthetic dataset.}
    }
     \vspace{-1em}
    \label{tab:xarm_synthetic_translation}
\end{table}

Tab.~\ref{tab:xarm_synthetic_rotation} and Tab.~\ref{tab:xarm_synthetic_translation} present the evaluation results of rotation and translation errors, respectively, where our method consistently outperforms previous approaches.
Even using a single view, our method surpasses the performance of the prior work~\cite{easyhec}, attributed to the precision of our pose initialization
Despite both EasyHeC~\cite{easyhec} and our method using the same differentiable-rendering-based optimization for camera pose resolution, our approach achieves superior accuracy.
This superiority is attributed to our method's utilization of the robot arm's kinematics model and a large pre-trained segmentation model to predict segmentation masks, which proves more accurate than the masks generated in~\cite{easyhec}.
Specifically, our method achieves a translation error of 0.135cm with 5 joint poses, outperforming EasyHeC~\cite{easyhec} which records a 0.206cm error under similar conditions.

\begin{table}[ht]
    \centering
    \renewcommand{\arraystretch}{1.2}
    \resizebox{\linewidth}{!}{
        \begin{tabular}{c|c|ccccc}
            \hline
            Method & \#views & \multicolumn{5}{c}{Rotation error ($^\circ$)} \\
            \Xhline{3\arrayrulewidth}
            Zhang et al.~\cite{zhang2023active} & 10 & \multicolumn{5}{c}{ 0.148} \\
            \hline
            Valassakis et al.~\cite{valassakis2022learning} & 1 & \multicolumn{5}{c}{4.4    } \\
            \hline
            Ours & 1 $\sim$ 5 & \textbf{0.685} & \textbf{0.204} & \textbf{0.145} & \textbf{0.130} & \textbf{0.117} \\
            \Xhline{3\arrayrulewidth}
        \end{tabular}
    }
    \caption{
        \textbf{Rotation error evaluation results on the xArm synthetic dataset.}
    }
    \vspace{-1em}
    \label{tab:xarm_synthetic_eih_rotation}
\end{table}
\begin{table}[ht]
    \centering
    \renewcommand{\arraystretch}{1.2}
    \resizebox{\linewidth}{!}{
                \begin{tabular}{c|c|ccccc} 
\hline
Method & \#views             & \multicolumn{5}{c}{Translation error (cm)}  \\ 
\Xhline{3\arrayrulewidth}
Zhang et al.~\cite{zhang2023active}      & 10                  & \multicolumn{5}{c}{ 0.315}           \\ 
\hline
Valassakis et al.~\cite{valassakis2022learning}     & 1                   & \multicolumn{5}{c}{1.340    }         \\ 
\hline
Ours   & 1 $\sim$ 5 & \textbf{0.503} & \textbf{0.179} & \textbf{0.128} & \textbf{0.117} & \textbf{0.112}       \\
\Xhline{3\arrayrulewidth}
\end{tabular}
    }
    \caption{
        \textbf{Translation error evaluation results on the xArm synthetic dataset.}
         }
    \vspace{-1em}
    \label{tab:xarm_synthetic_eih_translation}
\end{table}

\subsubsection{Eye-in-Hand Setting}\label{sec:eye-in-hand-evaluation}
Similar to the eye-to-hand setting, we evaluate our method on a synthetic dataset under the eye-in-hand setting and compare it to previous methods.
The evaluation involves 50 different camera poses, and we simulate the calibration process using SAPIEN~\cite{Xiang_2020_SAPIEN}.
For the approach proposed in Valassakis et al.~\cite{valassakis2022learning}, we synthesize 10000 images using a xArm7 robot arm and train their model.
For Zhang et al.~\cite{zhang2023active}, we reproduce their method with a chessboard of 4$\times$5 grids with a 5cm grid size and images captured at 10 different joint poses.
Compared to~\cite{valassakis2022learning}, which trains a neural network to regress the camera pose, our method not only requires no training but also achieves better accuracy.
Furthermore, in comparison to~\cite{zhang2023active}, our method achieves better accuracy with fewer joint poses.

\subsection{Evaluation on Real Dataset}

We conducted a comparative evaluation of our method against several previous approaches using the real-world Baxter dataset~\cite{lu2023image}.
This dataset comprises 100 images captured under the same camera pose but with 20 different joint poses.
The evaluation metrics include the percentage of correct keypoints (PCK) for both 2D and 3D, as presented in Tab.~\ref{tab:baxter_main_2d} and Tab.~\ref{tab:baxter_main_3d}, respectively.
The tables demonstrate that our method outperforms previous approaches in both 2D and 3D PCK, particularly under small thresholds.
For instance, when using 3 views, our method achieves a 2D PCK of 0.5 and 0.7 with 10 px and 20 px thresholds, respectively, compared to only 0.05 and 0.55 for EasyHeC~\cite{easyhec}.
Despite EasyHeC training a segmentation model on synthetic data with data augmentation, its generalization to real-world images remains challenging.
In contrast, our method leverages the generalization capability of the pre-trained image model to predict the segmentation mask. This not only eliminates the need for time-consuming and intricate data augmentation during training but also yields superior mask quality, resulting in higher PCK values.

\begin{table}[ht]
    \centering
    \renewcommand{\arraystretch}{1.15}
    \resizebox{\linewidth}{!}{
        \begin{tabular}{c|ccccccc}%
            \hline
            \multirow{2}{*}{Method} &
            \multicolumn{7}{c}{PCK 2D}\\
            \cline{2-8} & 10px & 20px   & 30px   & 40px   & 50px   & 100px  & 150px  \\ 
            \Xhline{3\arrayrulewidth}
            DREAM~\cite{lee2020dream} & - & 0.16 & 0.23 & 0.29 & 0.33 & 0.52 & 0.62 \\
            \hline
            OK~\cite{lu2022pose}   & - &  0.34 & 0.54 & 0.66 & 0.69 & 0.88 & 0.93 \\
            \hline
            IPE~\cite{lu2023image} (box)  & -  & - & - & - & 0.65 & 0.94 & 0.95 \\ %
            \hline
            IPE~\cite{lu2023image} (cylinder) & - & - & - & - & 0.80 & 0.91 & 0.93 \\ %
            \hline
            IPE~\cite{lu2023image} (CAD) & - &  - & - & - & 0.74 & 0.90 & 0.94 \\ %
            \hline
            EasyHeC (1view) & 0.1 & 0.35 & 0.55 & \textbf{0.75} & 0.90 & \textbf{0.95} & 0.95 \\ %
            \hline
            EasyHeC (2views) & 0.15 & 0.40 & 0.75 & \textbf{0.95} & \textbf{1.00} & \textbf{1.00} & \textbf{1.00} \\
            \hline
            EasyHeC (3views) & 0.05 & 0.55 & \textbf{0.85} & \textbf{1.00} & \textbf{1.00} & \textbf{1.00} & \textbf{1.00} \\
            \hline
            Ours (1view) & \textbf{0.25} & \textbf{0.5} & \textbf{0.75} & \textbf{0.75} & 0.85 & \textbf{0.95} & \textbf{1.00} \\
            \hline
            Ours (2views) & \textbf{0.5} & \textbf{0.55} & \textbf{0.9} & 0.9 & 0.9 & \textbf{1.00} & \textbf{1.00} \\
            \hline
            Ours (3views) & \textbf{0.5}  & \textbf{0.7} & \textbf{0.85} & 0.95 & \textbf{1.00} & \textbf{1.00} & \textbf{1.00} \\
            \Xhline{3\arrayrulewidth}
        \end{tabular}
    }

    \caption{
        \textbf{2D PCK evaluation results on the Baxter dataset.}
        2D PCK scores are given at different thresholds.
    }
    \vspace{-1em}
    \label{tab:baxter_main_2d}
\end{table}
\begin{table}[ht]
    \centering
    \renewcommand{\arraystretch}{1.15}
    \resizebox{\linewidth}{!}{
        \begin{tabular}{c|cccccc}
            \hline
            \multirow{2}{*}{Method} &
            \multicolumn{6}{c}{PCK 3D}\\
            \cline{2-7} & 2cm & 5cm & 10cm & 20cm & 30cm & 40cm \\ 
            \Xhline{3\arrayrulewidth}
            DREAM~\cite{lee2020dream} & 0.01 & 0.08 & 0.32 & 0.43 & 0.54 & 0.66 \\
            \hline
            OK~\cite{lu2022pose} & 0.10 & 0.34 & 0.54 & 0.66 & 0.69 & 0.88 \\
            \hline
            IPE~\cite{lu2023image} (box) & - & - & 0.8 & 0.95 & 0.95 & 0.95 \\
            \hline
            IPE~\cite{lu2023image} (cylinder) & - & - & 0.71 & 0.93 & 0.94 & 0.95 \\
            \hline
            IPE~\cite{lu2023image} (CAD) & - & - & 0.78 & 0.93 & 0.97 & \textbf{1.00} \\
            \hline
            EasyHeC (1view) & 0.10 & 0.65 & 0.90 & \textbf{1.00} & \textbf{1.00} & \textbf{1.00} \\
            \hline
            EasyHeC (2views) & 0.15 & \textbf{0.80} & \textbf{0.95} & \textbf{1.00} & \textbf{1.00} & \textbf{1.00} \\
            \hline
            EasyHeC (3views) & 0.15 & \textbf{0.80} & 0.90 & \textbf{1.00} & \textbf{1.00} & \textbf{1.00} \\
            \hline
            Ours (1view) & \textbf{0.15} & \textbf{0.75} & 0.95 & \textbf{1.00} & \textbf{1.00} & \textbf{1.00} \\
            \hline
            Ours (2views) & \textbf{0.2} & 0.6 & \textbf{0.95} & \textbf{1.00} & \textbf{1.00} & \textbf{1.00} \\
            \hline
            Ours (3views) & \textbf{0.3} & 0.65 & 0.90 & \textbf{1.00} & \textbf{1.00} & \textbf{1.00} \\
            \Xhline{3\arrayrulewidth}
        \end{tabular}
    }

    \caption{
        \textbf{3D PCK evaluation results on the Baxter dataset.}
    }
    \vspace{-2em}
    \label{tab:baxter_main_3d}
\end{table}
\begin{table}[ht]
    \centering
    \renewcommand{\arraystretch}{1.15}
    \resizebox{\linewidth}{!}{
        \begin{tabular}{c|cccc}
            \hline
            Method     & DREAM & EasyHeC & EasyHeC (SAM) & Ours         \\
            \Xhline{3\arrayrulewidth}
            Error (cm) & 1.5   & 0.4     & \textbf{0.3}  & \textbf{0.3} \\
            \Xhline{3\arrayrulewidth}
        \end{tabular}
    }
    \caption{
        \textbf{Real-world error high-precision targeting experiment under the eye-to-hand setting.}}
    \vspace{-1em}
    \label{tab:xarm_real_to_hand}
\end{table}
\begin{table}[ht]
    \centering
    \renewcommand{\arraystretch}{1.15}
    \resizebox{\linewidth}{!}{
        \begin{tabular}{c|ccc}
            \hline
            Method     & Zhang et al.~\cite{zhang2023active} & Valassakis et al.~\cite{valassakis2022learning}     & Ours \\
            \Xhline{3\arrayrulewidth}
            Error (cm) & 1.35                                & 4.30                                            & \textbf{0.31} \\
            \Xhline{3\arrayrulewidth}
        \end{tabular}
    }
    \caption{
        \textbf{Real-world error high-precision targeting experiment under the eye-in-hand setting.}
        Errors are computed across 10 tipping trails.
    }
    \vspace{-2em}
    \label{tab:xarm_real_in_hand}
\end{table}

\subsection{Real-world Evaluations}\label{sec:real-world-evaluation}
In addition to the real-world Baxter dataset~\cite{lu2023image}, we evaluated our method under a real-world setup.
Our experimentation involved using a xArm7 robot arm with a RealSense camera, positioned either on a nearby tripod or mounted on the end-effector for the eye-to-hand and eye-in-hand settings, respectively.
After applying our calibration method, we follow the procedure outlined in previous work~\cite{easyhec} for further evaluation.
This involves transforming the corner of an ArUco marker to the robot base coordinate system, tipping it, and manually measuring the error between the tip and corner.
The results are presented in Tab.~\ref{tab:xarm_real_to_hand} and Tab.~\ref{tab:xarm_real_in_hand}, where our method achieves the lowest error in both settings.
In the eye-to-hand setting, it is noteworthy that both EasyHeC (SAM) and our method utilize the SAM model for mask prediction.
However, they require manually annotated bounding-box prompts, while our method does not, enhancing its automation.

\subsection{Ablation Study}

\medskip\noindent\textbf{Automatic pose initialization.}
We conducted a comparison of different pose initialization methods between EasyHeC~\cite{easyhec} and our proposed approach.
The results are shown in Tab.~\ref{tab:pose_init_compare}.
EasyHeC trained a PVNet~\cite{peng2019pvnet} on synthetic data to initialize the camera pose at the robot arm's zero joint pose.
Both our sampling-based and feature-matching-based pose initialization achieve superior accuracy compared to PVNet.
The sampling-based initialization proves robust to mask quality, while the feature-matching-based method efficiently utilizes existing calibration results to initialize the pose.
Notably, the sampling-based method demonstrates higher accuracy but is more time-consuming, requiring the robot arm to be driven to multiple joint poses, taking approximately 10 minutes.
In contrast, the feature-matching-based method only requires the robot arm to have the same joint pose as the images in the database, avoiding the need for arm movement and proving more efficient.

\begin{table}[ht]
    \centering
    \renewcommand{\arraystretch}{1.15}
    \resizebox{\linewidth}{!}{
        \begin{tabular}{c|ccc}
            \hline
            Method                & Rot. error ($^\circ$) $\downarrow$ & Trans. error (cm) $\downarrow$ & Time $\downarrow$         \\
            \Xhline{3\arrayrulewidth}
            EasyHeC (PVNet)       & 3.94             & 2.40              & 10h+/100ms    \\
            \hline
            Ours (Sampling)       & 0.10             & 0.26              & 0/$\sim$10min \\
            \hline
            Ours (Feat. matching) & 2.33             & 1.80              & 0/6s          \\
            \Xhline{3\arrayrulewidth}
        \end{tabular}
    }

    \caption{
        \textbf {Ablation study for pose initialization.}
        Errors on the xArm eye-to-hand synthetic dataset.
        Training time/inference time are reported on an RTX 4090 GPU.
    }
    \label{tab:pose_init_compare}
\end{table}

\begin{figure}[ht]
    \centering
    \resizebox{1.0 \columnwidth}{!}{
        \begin{tabular}{ccc}
            \includegraphics[width=0.5\linewidth,trim={0cm 0cm 0cm 0cm},clip]{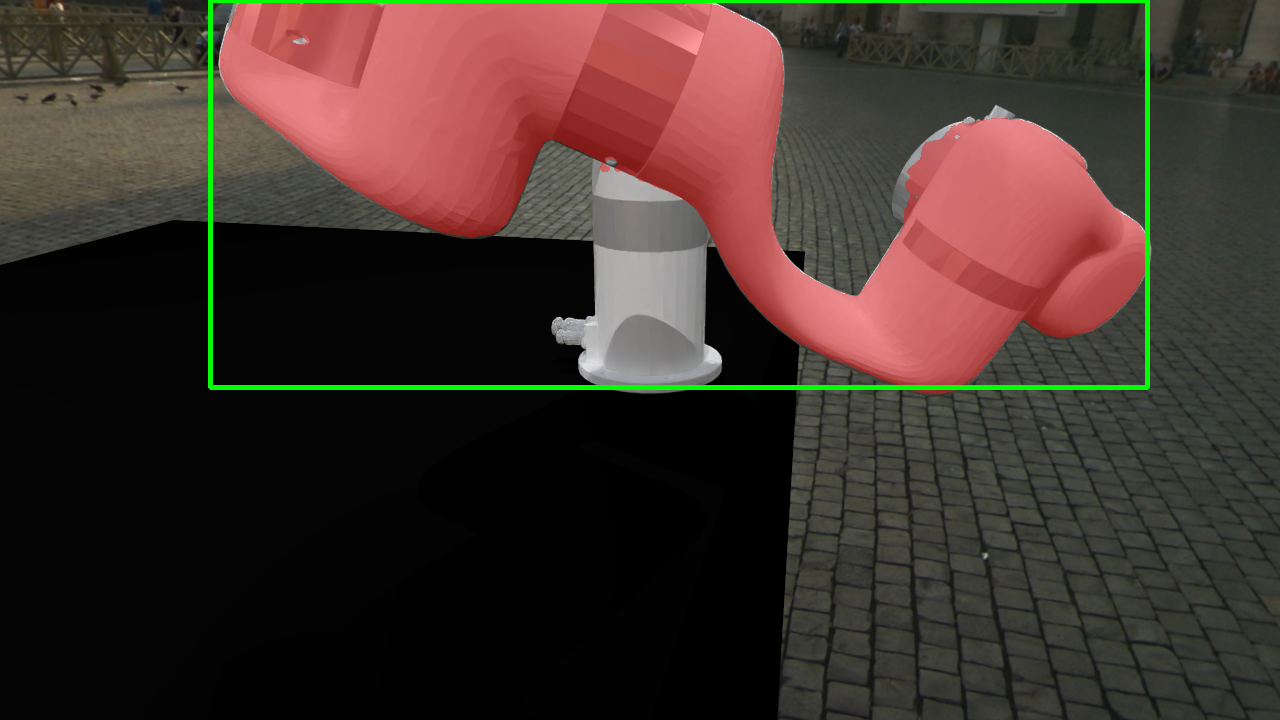} &
            \includegraphics[width=0.5\linewidth,trim={0cm 0cm 0cm 0cm},clip]{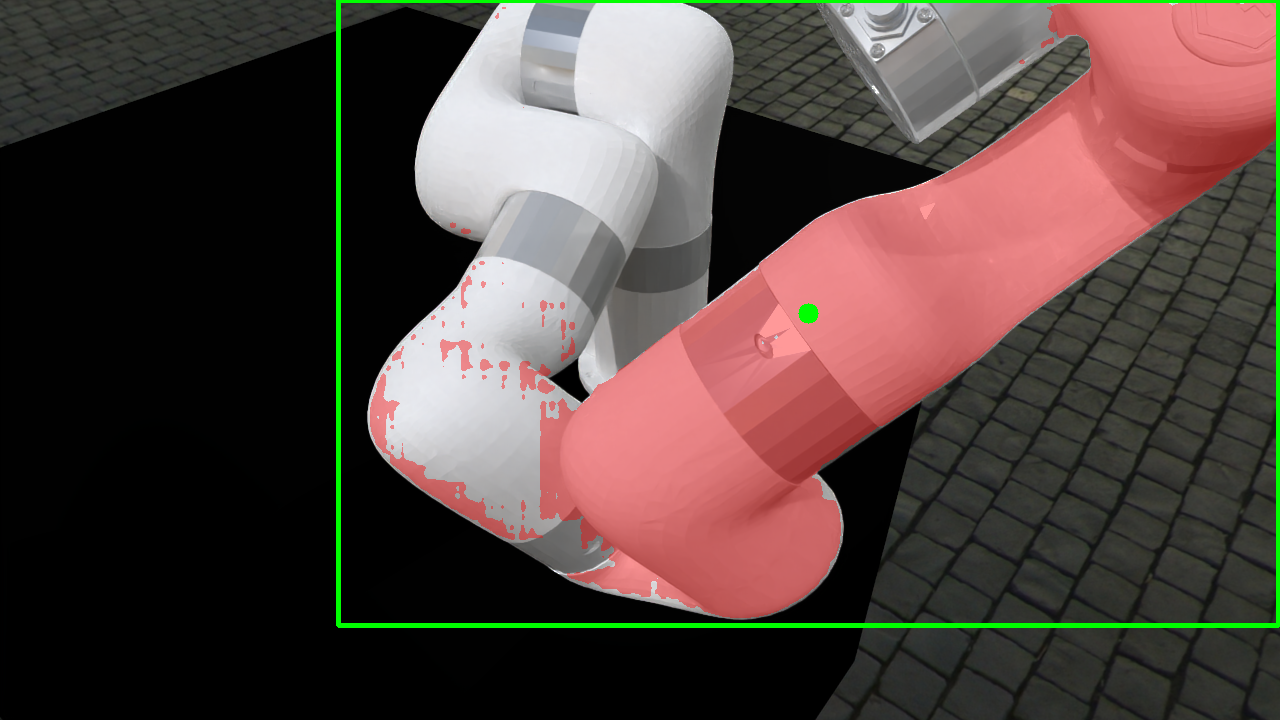} &
            \includegraphics[width=0.5\linewidth,trim={0cm 0cm 0cm 0cm},clip]{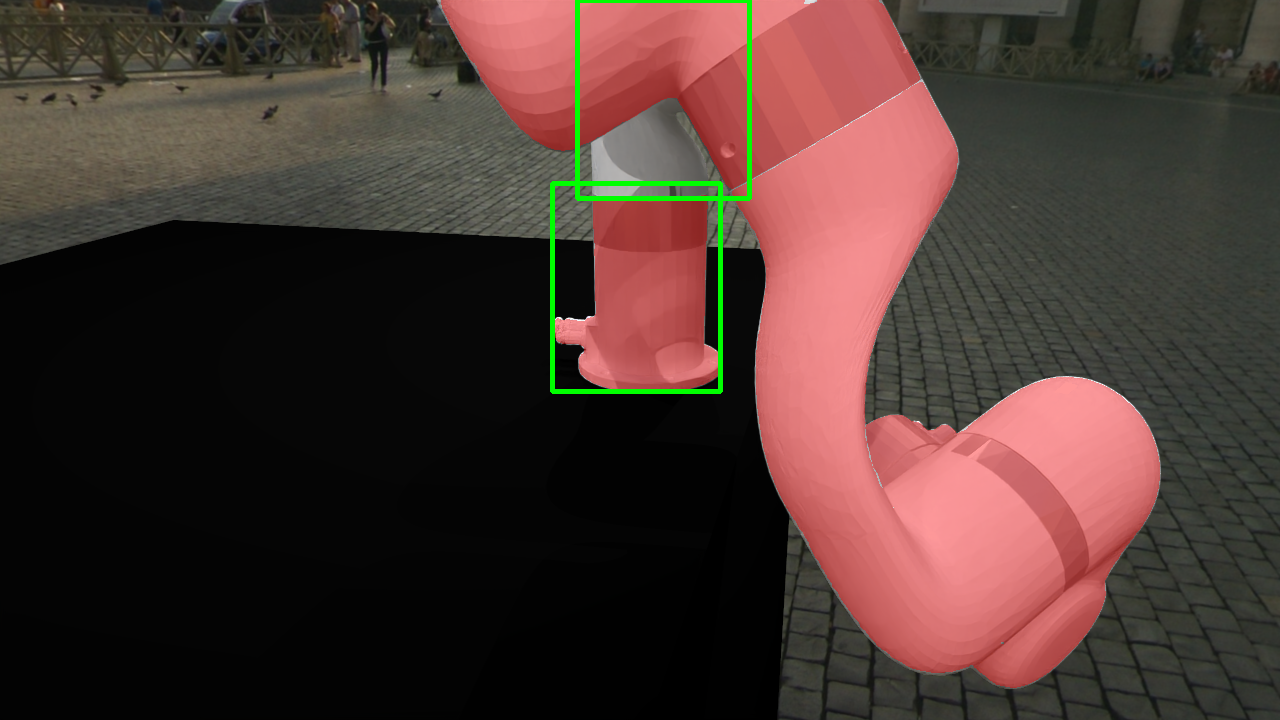} \\
            (a) & (b) & (c) \\
            \includegraphics[width=0.5\linewidth,trim={0cm 0cm 0cm 0cm},clip]{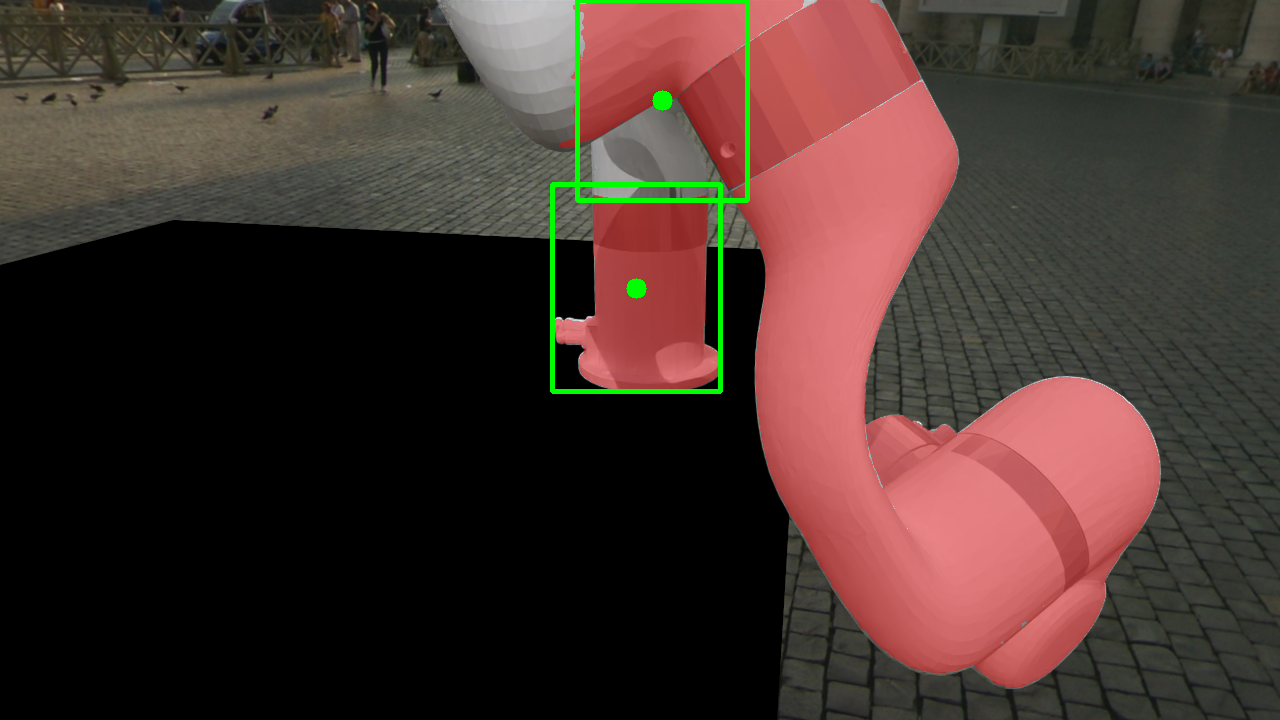} &
            \includegraphics[width=0.5\linewidth,trim={0cm 0cm 0cm 0cm},clip]{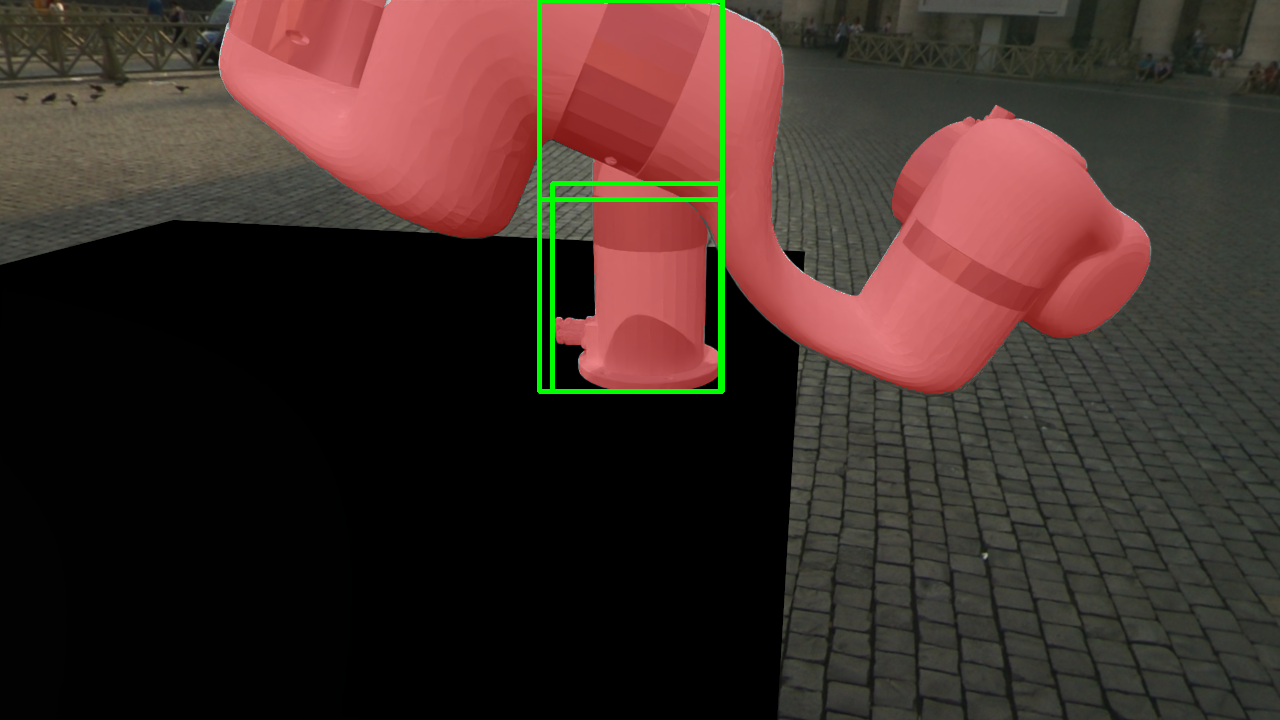} &
            \includegraphics[width=0.5\linewidth,trim={0cm 0cm 0cm 0cm},clip]{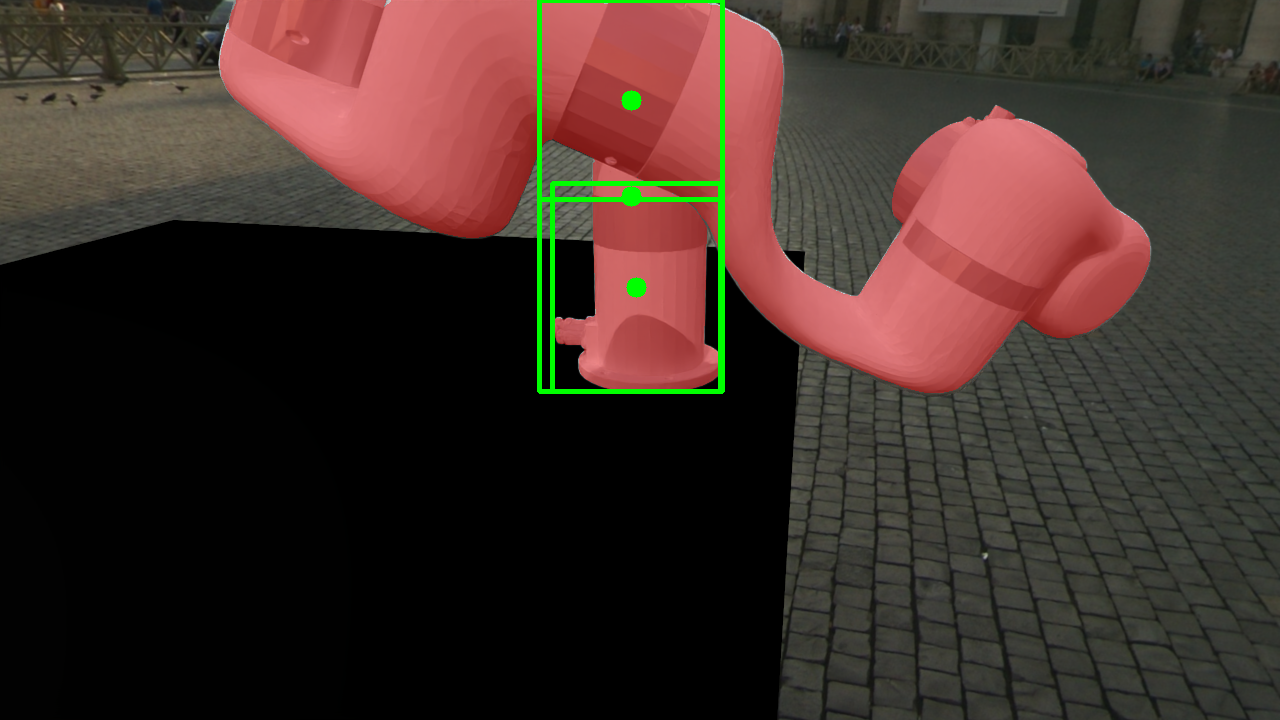} \\

            (d) & (e) & (f) \\
        \end{tabular}
    }
    \caption{\textbf{Ablation study on different types of prompts to the SAM model.}
    (a) A single bounding box of the whole robot arm as the prompt.
    (b) A single bounding box of the robot arm and its center point as the prompt.
    (c) Per-link bounding boxes as the prompt.
    (d) Per-link bounding boxes and their center points as the prompt.
    (e) Bounding boxes of each link and each connector as the prompt.
    (f) Bounding boxes of each link and each connector and their center points as the prompt.
    For clarity, we do not show all the prompts from (c) to (f).
    }
    \vspace{-1.5em}
    \label{fig:sam_ablation}
\end{figure}

\medskip\noindent\textbf{Different prompts to the SAM model.}\label{sec:ablation_study_sam}
In this ablation study, we compare different prompts to the SAM model on the xArm synthetic dataset.
The qualitative and quantitative results are shown in Fig.~\ref{fig:sam_ablation} and Tab.~\ref{tab:xarm_syn_mask_ablation}, respectively.
While the most straightforward prompt is to use a single bounding box for the entire robot arm, we observe that this prompt lacks accuracy, as shown in Fig.~\ref{fig:sam_ablation}(a).
This is because the space exploration module tends to produce a contorted joint pose, making it challenging for the SAM model to generate precise masks.
Additionally, real-world scenarios may involve occlusions from the table or other objects, as well as unwanted attachments on the robot arm (e.g., an in-hand camera), further diminishing the accuracy of the single bounding box prompt.
Combining the single bounding box prompt with a center point prompt can even lead to a reduction in accuracy, as depicted in Fig.~\ref{fig:sam_ablation}(b).
Although using per-link bounding boxes, as shown in Fig.~\ref{fig:sam_ablation}(c) and Fig.~\ref{fig:sam_ablation}(d), can enhance accuracy, it remains somewhat unstable, occasionally missing connectors between adjacent links.
The most accurate prompt involves using bounding boxes for each link and connectors between each pair of adjacent links, as demonstrated in Fig.~\ref{fig:sam_ablation}(e) and Fig.~\ref{fig:sam_ablation}(f). This comprehensive prompt configuration yields the highest accuracy in generating precise masks.

\begin{table}[ht]
    \centering
    \renewcommand{\arraystretch}{1.15}
    \resizebox{\linewidth}{!}{
        \begin{tabular}{c|ccccc}
            \hline
            Method   & Box prompt & Point prompt & Connector & Mask IoU $\uparrow$ \\
            \Xhline{3\arrayrulewidth}
            EasyHeC  & -          & -            & -         & 96.3                \\
            \hline
            Ours (a) & Single     & w/o          & w/o       & 92.3                \\
            \hline
            Ours (b) & Single     & w/           & w/o       & 93.0                \\
            \hline
            Ours (c) & per-link   & w/o          & w/o       & 93.7                \\
            \hline
            Ours (d) & per-link   & w/           & w/o       & 94.2                \\
            \hline
            Ours (e) & per-link   & w/o          & w/        & \textbf{98.2}       \\
            \hline
            Ours (f) & per-link   & w/           & w/        & \textbf{98.2}       \\
            \Xhline{3\arrayrulewidth}
        \end{tabular}
    }

    \caption{
        \textbf {Mask IoU comparison with different types of prompts to the SAM model tested on the xArm eye-to-hand synthetic dataset.}
        EasyHeC costs over 20 hours for training on an RTX 4090 GPU, while Ours requires no training.
        The indication of Roman numerals (a)-(f) are shown in Fig.~\ref{fig:sam_ablation}.
    }
    \label{tab:xarm_syn_mask_ablation}
\end{table}

    \section{Conclusion}
In this work, we proposed EasyHeC++, which can calibrate any robot arm in a marker-free, training-free, and fully automatic manner.
Our main approach consists of a pose initialization phase and a pose optimization phase.
By integrating the generalization ability of pretrained image models and the accuracy of optimization-based methods, EasyHeC++ achieves a fully automatic pipeline.
Experiments show that our method produces superior accuracy and degree of automation in both synthetic and real-world datasets for different robot arms and camera settings.
This work opens up more possibilities for lab and household applications~\cite{miller2004graspit,chen2023perceiving,jia2023chain, an2023rgbmanip,zobov2023auto,fang2023anygrasp,mahler2019learning} that require hand-eye calibration to reduce the sim-to-real gap, such as robot manipulation and grasping.

    \normalem
    \bibliography{root}

\end{document}